\documentclass[11pt, a4paper, logo, copyright]{googledeepmind}

\pdftrailerid{redacted}

\makeatletter
\renewcommand\bibentry[1]{\nocite{#1}{\frenchspacing\@nameuse{BR@r@#1\@extra@b@citeb}}}
\makeatother

\usepackage{kantlipsum, lipsum}
\usepackage{dsfont}
\usepackage{gdm-colors}
\usepackage[utf8]{inputenc} 
\usepackage[T1]{fontenc}    
\usepackage{hyperref}  
\usepackage{url}  
\usepackage{booktabs}  
\usepackage{multirow}  
\usepackage{amsfonts}  
\usepackage{nicefrac}  
\usepackage{microtype} 

\usepackage{soul}
\usepackage{subcaption}
\usepackage{tikz}
\usepackage{pgfplots}
\usepackage{wrapfig}
\usepackage{graphicx}
\usepackage{array}
\usepackage{algorithm} 
\usepackage[noEnd,indLines]{algpseudocodex} 
\usepackage{wrapfig}
\usepackage{cleveref}

\usepackage[dvipsnames]{xcolor} 
\usepackage[most]{tcolorbox}
\usepackage{caption}

\usepackage[authoryear, round]{natbib}

\graphicspath{{figures/}}

\newcommand{\ggefull}{GPT-2 Gradient Moment}
\newcommand{\gge}{GPT-2 GM}

\newcolumntype{H}{>{\setbox0=\hbox\bgroup}c<{\egroup}@{}}

\newcommand{\rawdata}[1]{\csname data-#1\endcsname}
\newcommand{\defdata}[2]{\expandafter\newcommand\csname data-#1\endcsname{#2}}

\newtcolorbox[auto counter,list inside=recipe,crefname={Recipe}{Recipes}]{recipebox}[2][]{%
colback=White!80!YellowGreen,colframe=ForestGreen,boxrule=2pt,arc=2mm,%
top=3mm,bottom=3mm,left=3mm,right=3mm,%
fonttitle=\bfseries,title=Recipe~\thetcbcounter: #2,#1}

\usepackage{amsmath,amsfonts,bm}

\def\eqref#1{equation~\ref{#1}}

\def\1{\bm{1}}

\DeclareMathAlphabet{\mathsfit}{\encodingdefault}{\sfdefault}{m}{sl}
\SetMathAlphabet{\mathsfit}{bold}{\encodingdefault}{\sfdefault}{bx}{n}

\correspondingauthor{(emielh,ruhe,jheek,mensink,salimans)@google.com}

\renewcommand{\today}{}

\author[1]{Emiel Hoogeboom}
\author[1]{David Ruhe}
\author[1]{Jonathan Heek}
\author[1]{Thomas Mensink}
\author[1]{Tim Salimans}

\affil[1]{Google DeepMind Amsterdam}

\title{Beyond Single Tokens: Distilling Discrete Diffusion Models via Discrete MMD}

\begin{abstract}
It is currently difficult to distill discrete diffusion models. In contrast, continuous diffusion literature has many distillation approaches methods that can reduce sampling steps to a handful.

Our method, Discrete Moment Matching Distillation (D-MMD), leverages ideas that have been highly successful in the continuous domain. Whereas previous discrete distillation methods \textit{collapse}, D-MMD maintains high quality and diversity (given sufficient sampling steps).
This is demonstrated on both text and image datasets.
Moreover, the newly distilled generators can \textit{outperform} their teachers.
\end{abstract}

\begin{document}
\maketitle

\begin{figure}[b]
    \centering
    \begin{minipage}[t]{.58\textwidth}\vspace{0pt}
    \includegraphics[width=0.99\linewidth]{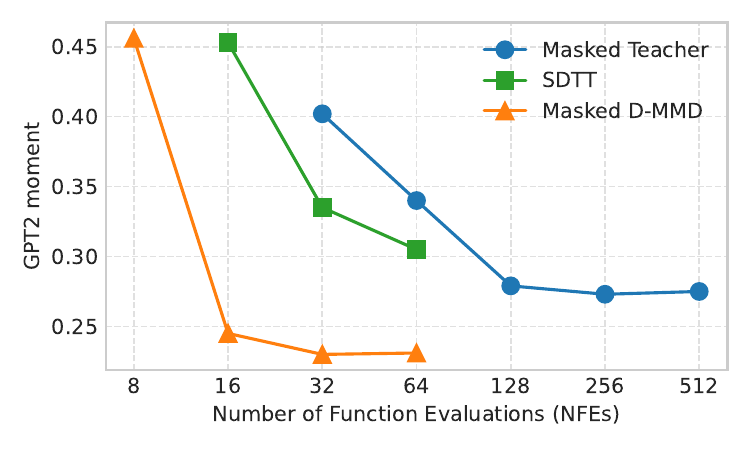}
    \end{minipage}
    \begin{minipage}[t]{.38\textwidth}\vspace{.8cm}
    \includegraphics[width=0.99\linewidth]{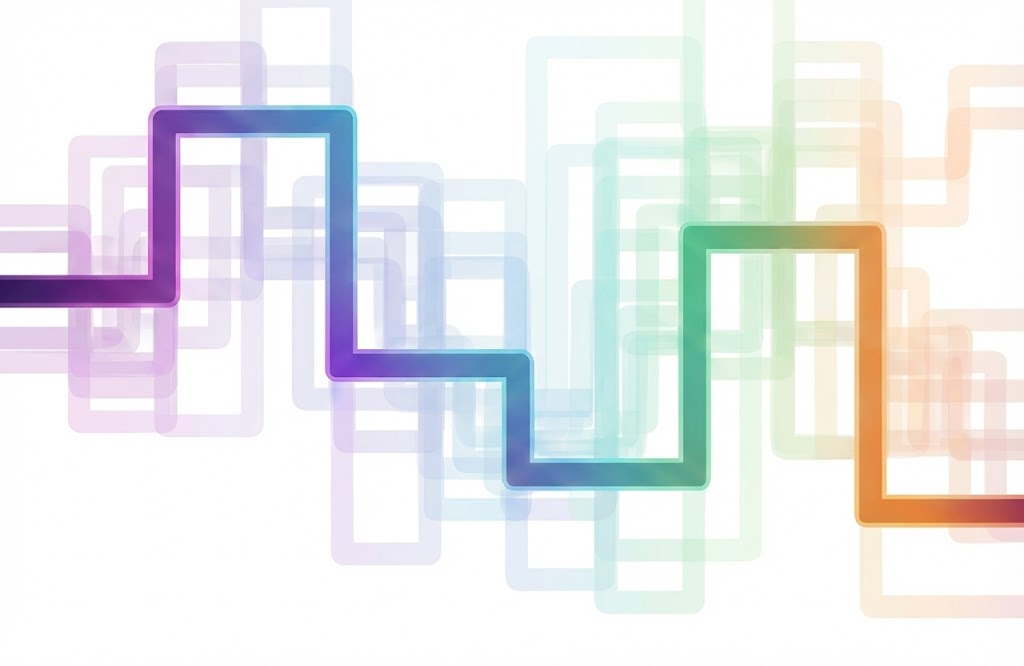}
    \end{minipage}
    \caption{D-MMD text generators match or outperform their teacher using fewer function evaluations.}
    \label{fig:overview}
\end{figure}

\section{Introduction}
Sampling from Discrete Diffusion Models requires many sampling steps. The probability of clean data given noisy data is modeled in a \textit{factorized} manner. I.e., each token is modeled independently conditioned on the previously generated tokens. 
As a result, errors from this assumed independence accumulate during the sampling iterations.

Discrete diffusion models perform forward passes on a block of tokens that they currently operate on. Whereas typical causal LLMs have under-utilization problems because they operate on single tokens, diffusion LLMs generally have high accelerator utilization. However, these models tend to need many iterations to converge to a reasonable generation, leading to high computing costs and strictly higher FLOPs. The fewer iterations one takes, the lower the cost.

In this paper we leverage insights from continuous diffusion to distill discrete diffusion models. 
Our paper generalizes the formulation of Moment Matching Distillation (MMD) \citep{salimans2024multistep} so that it can be used in more general settings. As our main focus is to distill discrete diffusion processes, we call this new algorithm \emph{Discrete-MMD} (D-MMD).
We show that one can distill few-step generators using D-MMD with sample quality surpassing their teachers on both text and image generation (see \autoref{fig:overview}).

\lstset{
    literate={“}{{``}}1
           {’}{{'}}1
           {”}{{''}}1,
    breaklines=true,
}

\section{Background}
\label{sec:background}

\paragraph{Diffusion Models}
Diffusion models are used to learn arbitrary distributions. Whereas continuous variables are often modeled with Gaussian noise diffusion, discrete variables are typically modeled with Uniform diffusion \citep{hoogeboom2021argmaxflows} or Masked diffusion \citep{austin2021structured}. The model learns to generate by approximating small sub-steps of the reverse process, often assuming dimensional independence. 

\begin{algorithm}[t]
\begin{algorithmic}
\Require Student generator $\hat{x}_\eta$, teacher model $\hat{x}_\theta$, auxiliary model $\hat{x}_\phi$, training step $i$, sampling steps $k$, dataset $\mathcal{D}$, weighting function $w(s)$, Loss function $L(\cdot, \cdot, \cdot)$.
\State $s, \delta_t\sim\mathcal{U}(0, 1), \mathcal{U}(0, \tfrac{1}{k})$
\State $t=\min(1, s+\delta_t)$

\State Sample from dataset $\mathcal{D}$ and diffuse to produce $z_t$.
\State Generate probability vector $\hat{x}_\eta(z_{t})$
\State Sample $x \sim \mathrm{Categorical}(p=\hat{x}_\eta(z_{t}))$.
\State Use sampler to draw sample $z_{s} | x, z_t$, for example using posterior $q(z_s|z_{t}, x)$, stopgrad on $z_{s}$. 
\If{$i$ is even} 
\State Minimize $\mathcal{L}_{\text{GEN}}(\eta) = L_s(\hat{x}_\eta(z_{t}), \hat{x}_\theta(z_s), z_s) - L_s(\hat{x}_\eta(z_{t}), \hat{x}_\phi(z_s), z_s) \text{ w.r.t. } \eta$.
\Else
  \State \textit{Optional} use soft target $x \leftarrow \hat{x}_\eta(z_t)$ (only possible for masked diffusion)
  \State Minimize $\mathcal{L}_\mathrm{AUX}(\phi) =  L_s(x, \hat{x}_\phi(z_s), z_s) + L_s(\hat{x}_{\theta}(z_s), \hat{x}_\phi(z_s), z_s)  \text{ w.r.t. } \phi$.
\EndIf
\caption{Discrete MMD Training}
\label{alg:mmd}
\end{algorithmic}
\end{algorithm}

Assume the data distribution $q(x)$ and diffusion process $z_t \sim q(z_t | x)$ for $t \in [0, 1]$, often implemented as a stochastic function $z_t = \operatorname{diffuse}(x, t)$. Optimizing diffusion models can be viewed as minimizing the following KL-divergence:
\begin{align}
\begin{split}
    \mathrm{KL} \left[ q(x) \big{|} p_\theta(x) \right] &\stackrel{c}{\leq} \mathrm{KL} \left[ q(z_0, \ldots, z_1 | x) \big{|} p_\theta(z_0, \ldots, z_1) \right] \\
    &\stackrel{c}{=} \mathbb{E}_{t \sim \mathcal{U}(0, 1)} \mathrm{KL} \left[q(z_{t-dt}|x, z_t) || p_\theta(z_{t-dt}|z_t) \right] dt^{-1}\,.
\end{split}
\label{eq:diffusion_elbo}
\end{align}
where the unknown entropy of the data $H_q(x)$ is omitted and constant with respect to $\theta$ and equality holds in the limit $dt \to 0$.

Many diffusion objectives simplify down to finding the conditional expectation under data, meaning that the optimal solution is:
\begin{align}
    \mathbb{E}_q[x|z_t] = \hat{x}_\theta(z_t), \text{ under } q(x, t, z_t) \,. \label{eq:diffusion_l_star}
\end{align}

In practice one does not have direct access to $\mathbb{E}_q[x|z_t]$. Therefore the objective is learned through a loss on samples drawn from a dataset referred to as $q(x, z_t, t)$, where $z_t$ is the diffusion of datapoint $x$.
A typical diffusion loss has the form:
\begin{align}
    \mathcal{L}(\theta) = \mathbb{E}_{q(x, t, z_t)}\left[w(t) L_t(x, \hat{x}_\theta(z_t), z_t)\right], \label{eq:diffusion_l}
\end{align}

In the continuous case, this formulation includes score matching \citep{song2021scorebasedsde} or \emph{probability flow} \citep{lipman2023flowmatching} in which case $L_t$ will simplify to a weighted squared error between $x$ and $\hat{x}_\theta(z_t)$: $L_t^{\mathrm{cont}}(x, \hat{x}_\theta(z_t), z_t) = \lVert x - \hat{x}_\theta(z_t) \rVert^2 \,.$
\paragraph{Discrete Diffusion}
In this work we consider both \emph{masked discrete diffusion} \citep{austin2021structured}, where the destruction process gradually transforms tokens into a special masking token, and  \emph{uniform diffusion} \citep{hoogeboom2021argmaxflows}, which transforms tokens into a uniform distribution.
Both of these can also be formulated as discrete flow matching \citep{gat2024discreteflowmatching}.

In particular, we have a discrete process that interpolates from data $x$ to a factorized stationary distribution $\pi$ so that $z_t \sim \mathrm{Cat}\left(\alpha_t x + (1 - \alpha_t) \pi\right)$, $t \in [0,1]$ and $\alpha_t$ is a suitable noise schedule. Let $\alpha_{t|s} = \alpha_t / \alpha_s$. The posterior of this process given $x, z_t$ equals \citep{Sahoo2024simpleandeffective}:
\begin{equation}
    q(z_s|z_t,x)=\text{Cat}\left(z_s| \frac{[\alpha_{t|s}z_t+(1-\alpha_{t|s})1\pi^\top z_t]\odot[\alpha_s x+(1-\alpha_s)\pi]}{\alpha_t z_t^\top x+(1-\alpha_t)z_t^\top\pi}\right)\,.
\end{equation}

While multiple losses can be considered, we limit ourselves to a simple weighted data cross-entropy loss of the form
\begin{align}
     L_t^{\mathrm{disc}}(x, \hat{x}_\theta(z_t), z_t) = \left[w(t) \, \mathrm{CE}\left(x| \hat{x}_\theta(z_t)\right) \right]\, \label{eq:discrete_l}
\end{align}
with the cross-entropy loss $\mathrm{CE}(x| \hat{x}) = -\sum_c x_c \log \hat{x}_c$.
In this case, the model still approximates $E_q[x | z_t]$.

\paragraph{Multistep Moment Matching}
In moment matching distillation \citep{salimans2024multistep} continuous diffusion models are distilled to few-step generators $g_\eta$ that can outperform their teacher diffusion models.
The MMD algorithm states that the conditional expectation of clean data should be identical between the data distribution $q$ and the sampling distribution $g_{\eta}$ of the distilled loss. 
The MMD loss is formulated as:
\begin{align}
    \mathcal{L}_{\text{MMD}}^\star(\eta) = \mathbb{E}_{g_\eta(x, t, z_t)} \left[w(t) \lVert \mathbb{E}_{q}[x|z_t] - \mathbb{E}_{g_\eta}[x|z_t] \rVert^2 \right]\,, \label{eq:mmd_l_star}
\end{align}
for which several approximations are made to realize a practical algorithm,
since the conditional expectation of the generator is not analytically available.
In their work the best performing method is an alternating optimization algorithm. 
The expectations are replaced by the output of the teacher model $\hat{x}_\theta$ and with the output of an auxiliary model $\hat{x}_\phi$. While the teacher model is fixed, the generator $g_\eta$ and the auxiliary model $\hat{x}_\phi$ are optimized with the objectives:
\begin{align}
    \mathcal{L}_{\text{MMD}}(\eta) &= \mathbb{E}_{g_\eta(z_t, s, z_s)}\left[w(s) \, \hat{x}_\eta(z_t)^\top \mathrm{sg}( \hat{x}_\phi(z_s) - \hat{x}_\theta(z_s)) \right]\,, \label{eq:mmd_alternating_1} \\ 
    \mathcal{L}_\mathrm{AUX}(\phi) &= \mathbb{E}_{g_\eta(z_t, s, z_s)}\left[w(s) \left( \lVert \hat{x}_\eta(z_t) - \hat{x}_\phi(z_s) \rVert^2 + \lVert  \hat{x}_\phi(z_s) - \hat{x}_\theta(z_s) \rVert^2\right)\right]\,,
    \label{eq:mmd_alternating_2}
\end{align}
\section{Discrete MMD: A generalization of MMD}
\label{sec:d_mmd}

Here we derive a more general form of the MMD equations that can be used in more general diffusion processes, such as discrete diffusion. A key observation is that the alternating optimization of Equations~\ref{eq:mmd_alternating_1} and \ref{eq:mmd_alternating_2} can be rewritten to a more general min-max formulation, neglecting constants:
\begin{align}\label{eq:d-mmd}
    \mathcal{L}_{\text{D-MMD}}(\eta) &= \min_\eta \max_\phi \mathbb{E}_{g_\eta(z_t, x, s, z_s)}\left[L_s(x, \hat{x}_\theta(z_s), z_s) -  L_s(x, \hat{x}_\phi(z_s), z_s) \textcolor{gray}{\, - \, L_s(\hat{x}_{\theta}(z_s), \hat{x}_\phi(z_s), z_s)} \right],
\end{align}
where the last term only regularizes the auxiliary model to remain close to the teacher, without changing the fixed-point of the algorithm.

In words, the generator $g_\eta$ aims to minimize the loss under the teacher while maximizing the loss under the auxiliary model. Simultaneously, the auxiliary model is trained to minimize the loss with the generator and is regularized to remain close to the teacher distribution.

\paragraph{Equivalence to continuous MMD}
To show that the D-MMD produces the same gradients as the MMD equations, recall that $L_s(x, \hat{x}, z_s) = w(s) ||x - \hat{x}||^2$. Let $x_{\eta} = \hat{x}_\eta(z_t)$ be shorthand notation,
\begin{equation}
    \nabla_\eta \mathcal{L}_{\text{D-MMD}}(\eta) = \nabla_\eta \left( L_s(x_{\eta}, \hat{x}_\theta(z_s), z_s) -  L_s(x_{\eta}, \hat{x}_\phi(z_s), z_s) \right) = 2\, w(s) \frac{d \hat{x}_\eta}{d\eta} \left( \hat{x}_\phi(z_s) - \hat{x}_\theta(z_s)\right),
\end{equation}
which is the same gradient as $2\nabla_\eta \mathcal{L}_{\text{MMD}}(\eta)$ assuming independence of $z_s$ on $\eta$ as done in \citet{salimans2024multistep}, by using a stop-gradient $\mathrm{sg(\cdot)}$. The equivalence for the loss of the auxiliary model 
follows directly from substitution of the loss terms and is not displayed here.

A fixed point for the algorithm occurs when $g_\eta$ generates exactly the teacher induced distribution. In this case the auxiliary model $\hat{x}_\phi(z_s)$ will equal the teacher $\hat{x}_{\theta}(z_s)$ and the loss will equal zero. 
In practice, the dynamics of adversarial optimization can be difficult and often depends on specific hyper-parameter settings.

\paragraph{Discrete D-MMD: matching probabilities}
The loss in Equation~\ref{eq:d-mmd} is difficult to optimize for discrete diffusion because there is no straightforward gradient from the categorical sample $x$ to $\eta$. Instead of drawing hard samples $x$, the soft probability vector $\hat{x}_\eta(z_t)$ is used, as also done in \cite{zhu2025dimo}. Simplifying the expression from the algorithm we observe that the equation is doing direct matching moments on expectation of $x$:
\begin{equation}
     \mathcal{L}_{\text{GEN}}(\eta) = \mathrm{CE}\left(\hat{x}_\eta| \hat{x}_\theta(z_s)\right) - \mathrm{CE}\left(\hat{x}_\eta| \hat{x}_\phi(z_s)\right) = -\sum_c (\hat{x}_\eta)_c \left( \log \hat{x}_\theta(z_s) - \log \hat{x}_\phi(z_s)\right)_c.
\end{equation}
Effectively, the gradient of this algorithm simply gives an update for $\hat{x}_\eta$ which is a delta of the log-probability of $\hat{x}_\theta(z_s)$ and $\hat{x}_\phi(z_s)$. Note that the update is still very similar to standard MMD. The main difference is that the update is now in log-probability instead of the standard output space. The loss for the auxiliary model is:
\begin{equation}
     \mathcal{L}_{\text{AUX}}(\phi) = \mathrm{CE}\left(x| \hat{x}_\phi(z_s)\right) + \mathrm{CE}\left(\hat{x}_\theta| \hat{x}_\phi(z_s)\right) = -\sum_c (x + \hat{x}_\theta(z_s))
     _c \log \hat{x}_\phi(z_s)_c.
\end{equation}
Here the auxiliary model is optimized to learn the expectation of the generator, $\mathbb{E}_{g_\eta(x, z_s)}[x|z_s] \stackrel{!}{=} \hat{x}_\phi(z_s)$, with a second regularization term that does not change the fixed point. For this algorithm the fixed point is $\mathbb{E}_{g_\eta(x, z_s)}[x|z_s] = \hat{x}_\phi(z_s) = \hat{x}_\theta(z_s)$. For masking diffusion and discrete flow matching, this algorithm can directly be used. For other types of diffusion where the optimal solution may not be $\hat{x}_\theta(z_t) = \mathbb{E}_q[x | z_t]$ (such as traditional uniform diffusion) we refer the readers to Appendix~\ref{app:other_diffusions}.

In \Cref{app:sufficiency_factorized_probs} we show that if the generator samples such that the teacher and auxiliary models match perfectly, we are guaranteed to sample according to the teacher distribution. Finally, an overview of the algorithm is given in \Cref{alg:mmd}.

\subsection{How can a factorized generator even learn correlated outputs?}
It may seem impossible that a factorized model is learning to correlate its outputs. However, the generator is a composition of two sampling steps. First, $\hat{x}_\eta(z_t)$ is a stochastic function that generates ``soft samples''. Subsequently a second step samples $x \sim \mathrm{Cat}(\hat{x}_\eta(z_t))$ hard tokens. Note that \textit{only the second step is factorized}.

Because the second step is factorized, the only way for the generator to minimize the moment matching loss is to \textit{correlate} the soft samples $\hat{x}_\eta(z_t)$ and reduce their \textit{output entropy}. This is not to be confused with the total entropy of the generator, because the sampling of soft tokens also contributes to its entropy. In practice, we observe that generators indeed reduce their output entropy to generate correlated outputs (see \autoref{tab:cifar10_noiseinput}).

\subsection{Correcting the bias of $\hat{x}_\eta$ for the auxiliary model.}
For training the auxiliary model, it is not always possible to use $\hat{x}_\eta$ as a soft target.
The reason is that $z_s \sim q(z_s | x, z_t)$ is a sample consistent with $x, z_t$, and not with $\hat{x}_\eta$ (despite $x \sim \mathrm{Cat}(\hat{x}_\eta)$). An exception is masked diffusion, because per dimension a masked $z_s$ does not provide information about $x$. For masked diffusion it is therefore equally valid to use either the soft $\hat{x}_\eta$ or the hard $x$. On the contrary for uniform diffusion the auxiliary model always needs to be trained on the hard samples.

\subsection{Temperature and top-p distillation}
In practice language models are often sampled using modified logits, for example through lower temperature sampling or top-p selection. This results in the samples being slightly more towards the mode of the distribution. 
Similar to the continuous MMD algorithm, where teacher guidance is incorporated during distillation to improve the image quality, we aim to distill student generators, which incorporate this teacher mode seeking in their sampling. 

For temperature distillation, the modification is relatively straightforward: the new teacher logits are computed as $s_\theta(z_s) = \tfrac{1}{\tau} \log \hat{x}_{\theta}(z_s)$ where $\tau$ is the temperature.

For top-p sampling, we need to be careful to avoid exploding gradients.
In top-p sampling, the idea is to select a subset of categories corresponding with a cumulative probability just over $p$ and mask out the other categories.
A typical top-p masking implementation takes in logits, and masks out the smallest categories with a very small value such as $-10^{20}$.
This however could lead to gradient spikes, as the teacher log-probability now is in the order of $-10^{20}$. Note that the softmax Jacobian of $\hat{x}_\eta$ is not sufficiently small to cancel this term out. Under this naive implementation, top-p distillation diverges in our experiments.
Instead of masking to $-10^{20}$, we found that it works to dynamically lower the logits by a constant: $s_\theta(z_s) \leftarrow s_\theta(z_s) - (1-\mathrm{mask}_{\mathrm{top-p}}) \cdot \Delta$, which roughly lowers the probability of the masked out categories by a factor $1/e^\Delta$, ignoring the correction effect on the softmax normalization term. In experiments we use $\Delta=2$, although the precise constant does not really matter,  as small log-probability differences will be discounted through the softmax Jacobian of $\hat{x}_\eta$ for low-probability events.

\section{Related work}
\label{sec:related_work}

\paragraph{Deterministic Diffusion Distillation}
The earliest distillations of diffusion models were \textit{deterministic}. These are based on the probability flow ODE, often approximated by the DDIM sampler. Early methods aimed to iteratively learn the trajectory using the iterative progressive distillation \citep{salimans2022progressive,meng2022ondistillation}. Later methods based on consistency models \citep{song2023consistency} use a more inductive approach where the generator is using itself as a target to solve for part of the trajectory \citep{kim2023consistency,song2023improvedconsistency,heek2024multistep,lu2024simplifying}.
Recently, flow-map or consistency-based distillation approaches have been applied to discrete data lifted to continuous space with standard diffusion models \citep{sahoo2025diffusion,roos2026categorical,lee2026one}. Currently, it remains to be seen whether these continuous models on discrete data can match the performance of discrete diffusion models. Furthermore, for both model classes it remains to be seen whether they can match the performance of standard autoregressive models.

\paragraph{Stochastic Diffusion Distillation}
Arguably a more successful method to distill diffusion models into generators is by \textit{stochastic distillation}, sometimes referred to as \textit{distribution matching} \citep{wang2023prolificdreamer,luo2024diff,yin2024one} which distill a diffusion model by approximately minimizing the KL divergence between the distilled generator and the teacher model. When the generator is single-step, MMD \citep{salimans2024multistep} is equivalent to the distribution matching approaches, but it tends to outperform them in few-step regimes.

\paragraph{Discrete Diffusion Models}
Direct concepts of continuous diffusion were adopted by \citep{sohldickstein2015diffusion,hoogeboom2021argmaxflows,austin2021structured} which pioneered diffusion on discrete data.
\citet{austin2021structured} proposed a generalized formulation and introduced an absorbing state or masked process.
\citet{chen2022analog} introduce Bit Diffusion, which applies continuous diffusion to the binary representations of discrete data.
More bridges between continuous and discrete diffusion were built by e.g. \citet{lou2023discrete}, who explored discrete versions of score matching and Tweedie's formula.
Arguably, masked diffusion became the leading paradigm in this research direction, with SOTA results achieved by e.g. MD4 \citet{shi2024simplified}.
Most recently, discrete diffusion can also be cast as a case of flow matching \cite{gat2024discreteflowmatching}.
While currently there still exists a performance gap between autoregressive and diffusion models, hybrid methods like 
\citet{arriola2025block} combine autoregressive and non-autoregressive techniques, also enabling variable-length generation.

\paragraph{Discrete Diffusion Distillation}
There have been a few distillation approaches that target discrete diffusion processes. SDTT \citep{deschenaux2025beyondautoregression} takes an approach reminiscent of progressive distillation but applied to discrete sampling. Although the approach tends to produce improvements to limited degree, it is fundamentally limited. For example, perfectly correlated coin tosses of two coins cannot be approximated with a single step of this approach. Due to the divergences chosen, SDTT will overcome the above mentioned limitation by directly dropping modes to achieve sampling speedups.

In Di4C \citep{hayakawa2024distillationdiscretediffusion} the shortcomings of factorized output distributions are recognized. The model outputs are extended to support mixture distributions, which allows the model to learn correlated outputs. Although effective to some degree, they tend to be limited in effect. One is often fighting an exponential of correlations between all tokens, and therefore the number of required mixtures also grows exponentially. In contrast, our D-MMD approach leaves the factorized output distribution unchanged. Instead, the generator can only match expectation moments if itself collapses the factorized output distribution. Another perspective is that our entire generator has become the mixture distribution.

In DiMO \citep{zhu2025dimo} it is shown how one can distill a single step generator from a masked diffusion model for image token generation. Although derived differently via straight-through softmax sampling, the resulting algorithm is equivalent to the implementation of D-MMD for the one-step case. Expanding on their approach, D-MMD generalizes to other types of processes (for example uniform diffusion) and supports few-step generators. These extensions make D-MMD applicable to a wider range of tasks such as high-quality text diffusion generators.

Concurrent to our work, IDLM \citep{li2026idlm} proposes a similar framework. The difference with IDLM is that the training algorithm generates the full $x$ and diffuses back to $z_t$, whereas our work samples from the posterior $q(z_s | z_t, x)$. We view this work as complementary.

\begin{figure}[]
    \centering
\begin{lstlisting}


“He’s in a really good spot. It’s the right situation, you don’t have to put him in, you shouldn’t be able to get him, and I think that is what I loved to do when he was young and didn’t put him in, which is how we did that,” Klineen said.

“He’s shown this year, with his growth in the system, he’s done a really, really, great job offensively. Over and over year he looks like he’s getting better. It’s definitely on the right path for him. It’s just early, right now, so we’ve got to get him and see how far it goes.

“I think he is definitely on the right path, I think he is playing on a high level. He’s made great strides, and he’s working hard. He’s got a lot of growth left in his body, he’s still growing. He has just got to get ready. We’ve got to get him and see if it goes well and then have him coming back next year, if it’s a long year. Hopefully it’s not. I want to get him ready, I just hope that he’s ready come next in. That’s the next step for him.”
\end{lstlisting}
    \caption{Excerpt of a random 1024-token sample generated using 16-step Masked D-MMD, not cherry picked.}
    \label{tab:placeholder}
\end{figure}

\section{Evaluating discrete diffusion models using Gradient Moments}
\label{sec:gradient_reference_error}

\begin{figure}[h]
    \centering
    \includegraphics[width=0.6\linewidth]{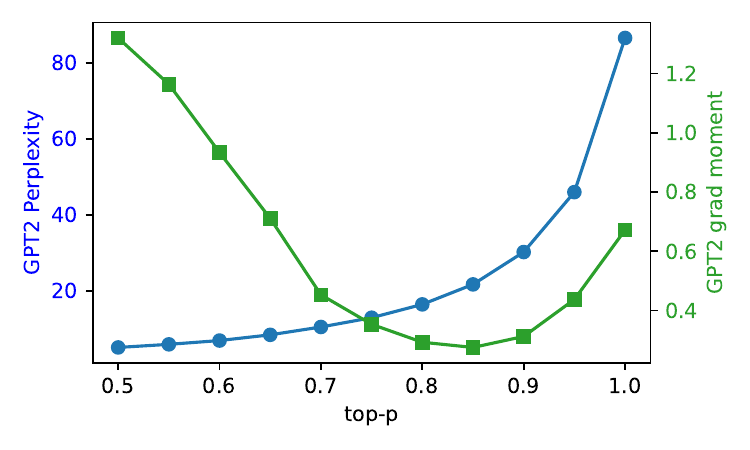}
    \caption{The perplexity metric keeps improving with lower temperature sampling while the grad moment eventually degrades.}
    \label{fig:gpt2_metric}
\end{figure}

Unlike standard autoregressive language models, distilled discrete diffusion models do not have a tractable sampling likelihood. This means that we cannot evaluate this model class with the standard perplexity metric. For this reason the literature often evaluates these models using \emph{generative perplexity}, where the samples from a discrete diffusion model are processed by an AR model like GPT-2 \citep{radford2019language}, and the perplexity of that AR model on the discrete diffusion samples is reported. The intuition is that samples are judged to be good if a reference LLM assigns a high probability to them. However, this is a flawed premise, as is also discussed in the literature \citep{azzopardi2003investigating,celikyilmaz2020evaluation}: high density samples are often not \emph{typical} \citep{meister2022typical},
meaning that they are not actually similar to the data. An example failure case of the generative perplexity metric is assigning a good score to ungrammatical generated samples that feature many repeated words. Fig.~\ref{fig:gpt2_metric} shows how perplexity and the grad moment metric are affected by top-p sampling. The grad moment eventually degrades when sampling at a low enough temperature.

Here we therefore propose a new metric for evaluating sample quality for discrete diffusion models, the \emph{Gradient Moment} of a reference model. The intuition behind this metric is that while the log-likelihood of a reference AR model on generated samples $\log p_{\theta}^{\text{LLM}}(x)$ is not indicative of sample quality, its \emph{gradient} $\nabla_{\theta} \log p_{\theta}^{\text{LLM}}(x)$ \emph{is}. If an AR model has been trained to convergence on a particular data distribution $q(x)$, its loss gradient on that distribution will be zero: $\mathbb{E}_{q(x)} \nabla_{\theta} \log p_{\theta}^{\text{LLM}}(x) = \mathbf{0}$. Conversely, if the loss-gradient of a trained LLM is large when evaluated on samples $x$, this means that $x$ does not look like the training data. We therefore propose to measure sample quality by the squared norm of this gradient. In practice, our reference LLM may have been trained on a different dataset than the distillation data, or training may not have fully converged: In that case, the loss-gradient evaluated on distillation data is not exactly zero. We therefore correct for this by centering the sample loss-gradient with respect to the data loss-gradient, resulting in the following evaluation metric:
\begin{equation}
\lVert \mathbb{E}_{g}[\nabla_{\theta} \log p_{\theta}^{\text{LLM}}(x)] - \mathbb{E}_{q}[\nabla_{\theta} \log p_{\theta}^{\text{LLM}}(x)] \rVert^2, \label{eq:gpt2moment}
\end{equation}
where $g$ represents our model's sampling distribution and $q$ is the data distribution. Although this Gradient Moment can be applied to any reference model, in remainder of the paper we choose GPT-2 \citep{radford2019language} as the reference model. The resulting metric is thus the \emph{\ggefull} (\gge). When our sampling distribution is identical to the training data, $g = q$, the metric will attain its lowest possible value of zero. This means that the reference model is unable to distinguish our samples from the ground truth, in the sense that it would not update its parameters when finetuning on our generated data. If the model \emph{is} able to distinguish our samples, the metric will be larger than zero.

In practice we calculate an unbiased stochastic approximation of \eqref{eq:gpt2moment} by calculating gradients on two independent minibatches at a time and taking their inner product.
\begin{equation}
(\nabla_{\theta} \log p_{\theta}^{\text{LLM}}(x^{g}_{1}) - \nabla_{\theta} \log p_{\theta}^{\text{LLM}}(x^{q}_{1}))^{T}(\nabla_{\theta} \log p_{\theta}^{\text{LLM}}(x^{g}_{2}) - \nabla_{\theta} \log p_{\theta}^{\text{LLM}}(x^{q}_{2})), \label{eq:batchgpt2moment}
\end{equation}
where $x^{g}_{1}, x^{g}_{2}$ represent independent batches of samples from our model, and $x^{q}_{1}, x^{q}_{2}$ are independent batches of training data. This stochastic approximation can then be averaged over many batches in order to get a low variance estimate of the quality of our model. This is similar to the loss proposed by \cite{salimans2024multistep} for distilling (continuous) diffusion models, but here we use it as a metric to compare a model to the data distribution, using a reference model as judge.

Although our experiments in this paper are focused on \emph{unconditional generation}, an advantage of the proposed metric is that it is equally valid when conditioning our samples on a prompt or other prefix $x_{c}$: In that case we simply use conditional likelihoods of the form $p_{\theta}^{\text{LLM}}(x|x_{c})$ in the equations above. This is a meaningful advantage of the reference model gradient norm compared to other sampling based methods such as FID \citep{heusel2017gans}.

\section{Experiments}
\label{sec:experiments}
In this section we show that D-MMD can distill discrete diffusion teachers very effectively. Because D-MMD is a stochastic distillation method, we have to rely on metrics that match distributions to study how successful the distillation is. For images we rely on the FID metric, whereas for text we rely on \gge~(see \autoref{sec:gradient_reference_error}).\footnote{The original time of writing of this paper was September 2025.}
\clearpage  

\subsection{CIFAR-10}
\begin{table}[t]
    \centering
    \begin{tabular}{l|rrrrrrrrr}
        \toprule
         \bfseries Model & 4 & 8 & 16 & 32 & 64 & 128 & 256 & 512 & 1024 \\ \midrule
        Uniform Teacher & & & 36.3 & 17.1 & 10.7 & 8.6 & 7.9 & 7.6 & 7.5 \\
        Uniform D-MMD & 7.1 & 5.0 & 4.1 & \textbf{3.7} & 3.8  \\ \midrule 
        Masked Teacher & & & 122.9 & 47.1 & 20.0 & 11.1 & 7.8 & 6.7 & 6.4 \\
        Masked D-MMD & 22.3 & 12.7 & 5.3 & 3.8 & \textbf{3.5} \\
        \bottomrule
    \end{tabular}
    \caption{Overview of D-MMD generators and base model sample quality for different NFEs on CIFAR10.
    Results are measured in FID with 50K samples compared to the train dataset. D-MMD models \textit{substantially} outperform their teacher while using a fraction of the NFEs. 
    }
    \label{tab:cifar10_overview}
\end{table}

In this first set of experiments we train diffusion models to generate unconditional images.
The models are trained on the 32x32x3 images in the CIFAR10 dataset.
We train a model directly on the $\{0, \ldots, 255\}^{32\times32\times3}$ pixel values, resulting in a total of 3072 tokens that need to be generated. 
We evaluate the performance using the FID metric, which notwithstanding the flaws, is still one of the better metrics to measure distances between distributions of (generated) images. 

On this dataset we train a masked and uniform diffusion model. These models tend to perform worse than standard diffusion models because there is no inductive bias: every pixel value is a unique token in the vocabulary. 
The uniform diffusion teacher achieves an FID of 7.5 and the masked diffusion teacher an FID of 6.4 using 1024 denoising steps.\footnote{Note: continuous (standard) diffusion models easily obtain an FID of around 3 \citep{ho2020denoising}.}

Impressively, D-MMD is able to distill much better generators at only a fraction of the denoising steps compared to the original teacher (\autoref{tab:cifar10_overview}). 
For uniform diffusion models an FID of 3.7 is achieved in 32 steps versus an FID of 7.5 for a 1024-step teacher. For Masked diffusion models, the distilled generator outperforms the teacher with 16 steps, and obtains an FID of 3.5 with only 64 uniform denoising steps.
In conclusion, both uniform and masked D-MMD achieve a substantially better Pareto front of steps vs FID than their teachers.

\begin{table}[t]
    \centering
    \begin{tabular}{l|rrrrrrrrr}
        \toprule
         \bfseries Model & 8 & 16 & 32 & 64 & 128 & 256 & 512 \\\midrule
        Uniform Teacher ($p=0.50$) & & & 0.375 & 0.326 & 0.330 & 0.324 & 0.313  \\
        Uniform D-MMD ($p=0.70/0.70$) & 0.337 & 0.310 & 0.307 & 0.316 \\ \midrule
        Masked Teacher ($p=0.85$) & & & 0.402 & 0.307 & 0.297 & 0.275 & 0.275 & \\
        Masked D-MMD ($p=0.85$) & 0.456 & 0.236 & \textbf{0.225} & \textbf{0.231} \\ \midrule
        AR Baseline & \multicolumn{8}{c}{0.061} \\
        \bottomrule
    \end{tabular}
    \caption{Text D-MMD generators and base model sample quality for different NFEs measured in \gge~ (see \autoref{sec:gradient_reference_error}).
    D-MMD models outperform their teacher at a fraction of the NFEs.}
    \label{tab:text_mmd_overview}
\end{table}

\subsection{Text}

For text generation we train on Open Web Text (OWT) and take the last $2\%$ as a validation set. Because generative perplexity can be gamed by lower temperature sampling (either intentionally or unintentionally through biased samplers), we use the \gge~ metric to measure distance from the distribution. 

Similar to image experiments, we train masked and uniform diffusion teacher models and measure their performance by generating 1024 tokens unconditionally using increasing number of denoising steps. We tune the top-p value for the best \gge. The results are in \autoref{tab:text_mmd_overview}. 
The Masked D-MMD generator already outperforms the teacher using only 16 steps, achieving 0.236 \gge.
Similar to the results for images, both the uniform and masked generators consistently outperform their teacher counterparts and improve the whole Pareto front.

\subsection{Block autoregressive diffusion}

\begin{table}[t]
    \centering
    \begin{tabular}{l|rrrrrrrrr}
        \toprule
         \bfseries Model & 16 & 256 \\\midrule
        256-Block Uniform Teacher ($p=0.9$) & - & 0.225  \\
        256-Block Uniform D-MMD ($p=0.7$) & 0.225 & - \\
        \bottomrule
    \end{tabular}
    \caption{Block auto-regressive diffusion model with block size 256. 16-step D-MMD matches the performance of the 256-step teacher.}
    \label{tab:text_mmd_block_ar}
\end{table}

Rather than generating an entire sequence at once, a more realistic setup would be to use a diffusion model to generate a limited block of tokens conditioned on an auto-regressive encoder. This combines the training efficiency and efficient inference of an AR model with the parallel sampling of diffusion. In this experiment, the 16-step D-MMD generator matches the performance of the 256-step teacher (see \autoref{tab:text_mmd_block_ar}).

\begin{table}[t]
    \centering
    \begin{tabular}{l|rrrrrrrrr}
        \toprule
         \bfseries Method & NFE & FID \\\midrule
         Di4C Teacher & 40 & 8.0 \\
         Di4C (hybrid) & 20 & 9.5 \\
         Di4C & 10 & 20.6 \\
         \midrule
        Uniform Teacher & 512 & 7.6 \\
         & 64 & 10.7 \\
        Uniform D-MMD & 8 & 5.0 \\
        & 16 & 4.1 \\
        & 32 & 3.7 \\ \midrule
        Masked Teacher & 512 & 6.7 \\
         & 64 & 20.0 \\
        Masked D-MMD & 16 & 5.3 & \\
        & 32 & 3.8 & \\
        & 64 & \textbf{3.5} \\
        \bottomrule
    \end{tabular}
    \caption{Comparison with literature on CIFAR10. D-MMD considerably outperforms existing methods using fewer NFEs.}
    \label{tab:cifar10_related}
\end{table}

\begin{table}[t]
    \centering
    \begin{tabular}{l|rrrrrrrrr}
        \toprule
         \bfseries Method & NFE & \gge~$\downarrow$ & GPT2 Perplexity $\downarrow$ & Sample entropy \\\midrule
        Duo + DCD & 4 & & 108.2 & 4.82 \\
        Duo + Di4C & 4 & & 150.7 & 4.81 \\
        MDLM + SDTT & 4 & & 339.7 & 5.38 \\
        MDLM + Di4C & 4 & & 239.3 & 5.40 \\
        FMLM & 4 & & 76.4 & 5.05 \\ \midrule
        Masked Teacher  & 256 & 0.275 & 22.5 & 5.13 \\
         & 128 & 0.295 & 23.9 & 5.17 \\
         & 64 & 0.307 & 26.0 & 5.19 \\
        SDTT (reimpl.) & 64 & 0.293 & 26.9 & 5.17 \\
        & 32 & 0.340  & 30.4 & 5.18 \\
        Masked D-MMD 
         & 4 & 0.820 & 20.3 & 4.60\\
         & 16 & 0.236 & \textbf{17.2} & 5.00 \\
        & 32 & \textbf{0.225} & 19.4 & 5.05 \\ \midrule
        Data & & 0.000 & 15.4 & 5.44 \\ \midrule
        \textcolor{gray}{Masked Teacher} & \textcolor{gray}{256} & \textcolor{gray}{0.672} & \textcolor{gray}{85.9} & \textcolor{gray}{5.59} \\
        \textcolor{gray}{($p=1.0$)} & \textcolor{gray}{128} & \textcolor{gray}{0.711} & \textcolor{gray}{91.1} & \textcolor{gray}{5.61} \\
        & \textcolor{gray}{64} & \textcolor{gray}{0.781} & \textcolor{gray}{101.0} & \textcolor{gray}{5.63} \\
        \textcolor{gray}{Masked D-MMD} & \textcolor{gray}{4} & \textcolor{gray}{0.719} & \textcolor{gray}{66.1} & \textcolor{gray}{5.44} \\
        \textcolor{gray}{($p=1.0$)} & \textcolor{gray}{16} & \textcolor{gray}{0.558} & \textcolor{gray}{67.7} & \textcolor{gray}{5.57} \\
        & \textcolor{gray}{32} & \textcolor{gray}{0.578} & \textcolor{gray}{72.1} & \textcolor{gray}{5.57} \\
        \bottomrule
        
    \end{tabular}
    \caption{Comparison with literature on OWT measured in \gge~(lower is better), generative perplexity (should not be too high) and sample entropy (should not be too low). D-MMD is able to achieve even better results in fewer steps.}
    \label{tab:text_related}
\end{table}

\begin{table}[t]
    \centering
    \begin{tabular}{ll|rrrrrrrrr}
        \toprule
         \bfseries D-MMD Masked & & 4 & 8 & 16 & 32 & 64 \\ \midrule
        without noise & (FID) & 151\textcolor{white}{.0} & 37.0 & 14.7 & 7.7 & 6.0 \\
        \hspace{.5cm} & (generator output entropy) & 1.26 & 1.37 & 1.57 & 1.86 & 1.91 \\ \midrule
        with noise & (FID) & 22.3 & 12.7 & 5.3 & 3.8 & \textbf{3.5} \\
        \hspace{.5cm} &(generator output entropy) & 1.01 & 1.29 & 1.53 & 1.76 & 1.83 \\
        \bottomrule
    \end{tabular}
    \caption{Noise input conditioning is important for masked distillation. Fewer steps require more generator output collapse, and generators with noise conditioning can collapse their factorized output distribution more.
    }
    \label{tab:cifar10_noiseinput}
\end{table}

\subsection{Comparison related work}
In this section we compare to the discrete diffusion distillation literature. For Di4C, results in the main paper are available on CIFAR10. Note that Di4C is actually at an advantage here, because its teacher model is trained using a discrete process that mimics the destruction of a Gaussian process. As a result, Di4C is able to achieve a teacher FID of 8.0 using only 40 steps. Nevertheless, because D-MMD outperforms the teacher models it still outperforms Di4C with 5.0 using only 8 steps with the uniform generator (see \autoref{tab:cifar10_related}). 

Recall that a metric such as generative perplexity is roughly measuring your distance from a mode, and collapsed models can easily score generative perplexities near 1.0\footnote{For example the sentence "hahahahahahaha" repeated also has a perplexity near 1.0} (the optimum). Instead, we measure performance with \ggefull~(\gge), which is somewhat more robust to this. Here we do see that even though SDTT improves upon the teacher model, it still degrades over repeated distillation rounds and is outperformed by D-MMD (see \autoref{tab:text_related}). Especially the \gge~metric highlights this degradation. The optimal top-p was chosen at $p=0.85$ by sweeping, measuring \gge~on the masked teacher. SDTT (reimpl.) and D-MMD use the same teacher. For completeness, we also show results without top-p $p = 1.0$. For other related works, \citep{sahoo2025diffusion,roos2026categorical} the results were taken from \citep{lee2026one}.

\subsection{Conditioning the generator on input noise}
In theory the generator should have access to a noise source to be able to generate a distribution. However, in \citet{salimans2024multistep} it was noted that in practice no input noise is required for Gaussian diffusion distillation. However, in the case of 1-step masked generation \citep{zhu2025dimo} noise conditioning turned out to be important. For images we learn a projection of a 2D Gaussian noise pyramid to be added to the residual. For text we learn a projection of plain Gaussian noise.

In our case we find that masked distillation performs much better with an extra noise source (see \autoref{tab:cifar10_noiseinput}). In that case, the generator is able to collapse its output distribution more and achieves much better sample quality. In contrast, for uniform diffusion we did not observe any meaningful improvements. As is the case with Gaussian diffusion, for uniform diffusion there may already be sufficient noise in $z_t$ that the generator is able to use. All other masked distillation experiments in this paper condition on input noise.

\subsection{Discussion on students outperforming teachers}
It may seem counterintuitive that students can outperform their teachers. However, teachers are trained using maximum likelihood which is known to be mode-covering. Mode-collapsing behavior is often induced by reducing temperature or top-p sampling. 

Many distillation approaches such as D-MMD have an adversarial component and generate samples based on the student, which both are reminiscent of reverse-KL optimization. D-MMD may move more density towards modes without fully collapsing, which is typically desired for samples from an image or language generator.

A paradoxical side-effect is the following: suppose the student is better than the teacher for a certain number of steps. Then, the student's performance will degrade at some point even as sampling steps increase, as that performance will converge to the teacher's at high step counts.

\section{Conclusions}
In summary, D-MMD is a new technique that allows for a principled way to distill discrete diffusion processes into few-step generators. In experiments, generators tend to outperform their teachers considerably, using only a fraction of the denoising steps.

\clearpage
\bibliography{main.bib}
\bibliographystyle{abbrvnat}
\clearpage
\appendix
\clearpage
\section{Sufficiency of matching first moments}
\label{app:sufficiency_first_moments}
Let $q(x, z_0, \dots, z_1)$ be a diffusion process.
In the following we give a simple argument motivating why the first moment criterion
\begin{align}
    \mathbb{E}_{p_\eta}[x|z_t] \stackrel{!}{=} \mathbb{E}_q[x|z_t] \label{eq:exp_eq}
\end{align}
for all $t \in [0, 1]$ leads to
\begin{align}
    p_\eta(x) = q(x)\,.
\end{align}

Let $q(z_{t-dt}|z_t, x)$ be the analytically available ground-truth posterior of the forward diffusion process and 
\begin{align}
\hat{q}(z_{t-dt}|z_t) = q(z_{t-dt}|z_t, \mathbb{E}_q[x|z_t])\,.
\end{align}
For $\lim_{dt \to 0}$, we have
\begin{align}
    q(x, z_0, \dots, z_1) = q(z_1) \prod \hat{q}(z_{t-dt}|z_t)\,.
\end{align}
This holds because $q(z_s|z_t, x)$ is linear in $x$ in the $dt \to 0$ limit, so that $\mathbb{E}_{q(x|z_t)}[q(z_s|z_t, x)] = q(z_s|z_t, \mathbb{E}_q[x|z_t]) = \hat{q}(z_s|z_t)$.
Using $p_\eta(z_1) = q(z_1) = \mathrm{Categorical}(\pi)$ and the equality of conditional expectations, we immediately have
\begin{align}
p_\eta(x, z_0, \dots, z_1) = q(x, z_0, \dots, z_1)\,.
\end{align}

By marginalization, the result follows.

\section{Sufficiency of matching factorized probabilities}
\label{app:sufficiency_factorized_probs}
We can construct a similar argument as before.
Let
\begin{align}
    \hat{q}(z_{t-dt}|z_t) = \mathbb{E}_{\hat{q}(x|z_t)}\left[q(z_s|z_t, x) \right]
\end{align}
where 
\begin{align}
    \hat{q}(x|z_t) = \prod_{d=1}^D q(x_d|z_t)
\end{align}
is the (factorized) product of the true posterior $q(x|z)$ marginals $q(x_i|z_t)$.
Then it can be shown (e.g. \citet{gat2024discreteflowmatching}) that for $\lim_{dt \to 0}$ 
\begin{align}
    q(x, z_0, \dots, z_1) = q(z_1) \prod \hat{q}(z_{t-dt}|z_t) \,.
\end{align}
It follows that if we have matching priors, the generator $p_\eta$ should only sample such that the factorized distributions match for all $t \in [0, 1]$ to guarantee $p_\eta(x)=q(x)$.

\clearpage
\section{Extended Results: CIFAR10}
In this section we provide more detailed results for the main results presented in the paper.

\begin{figure*}[t]
    \begin{subfigure}[b]{.475\textwidth}
        \centering
        \includegraphics[width=\textwidth]{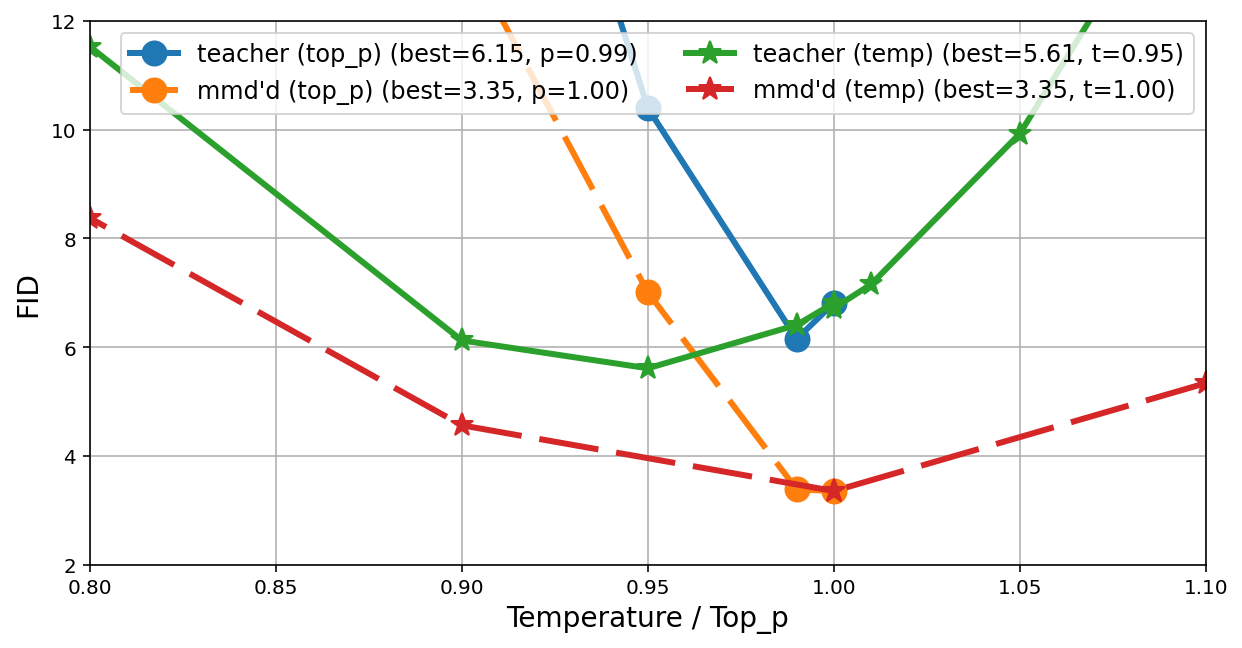}
        \caption{Masked diffusion, temperature and top $p$ sampling.}
    \end{subfigure}
    \begin{subfigure}[b]{.475\textwidth}
        \centering
        \includegraphics[width=\textwidth]{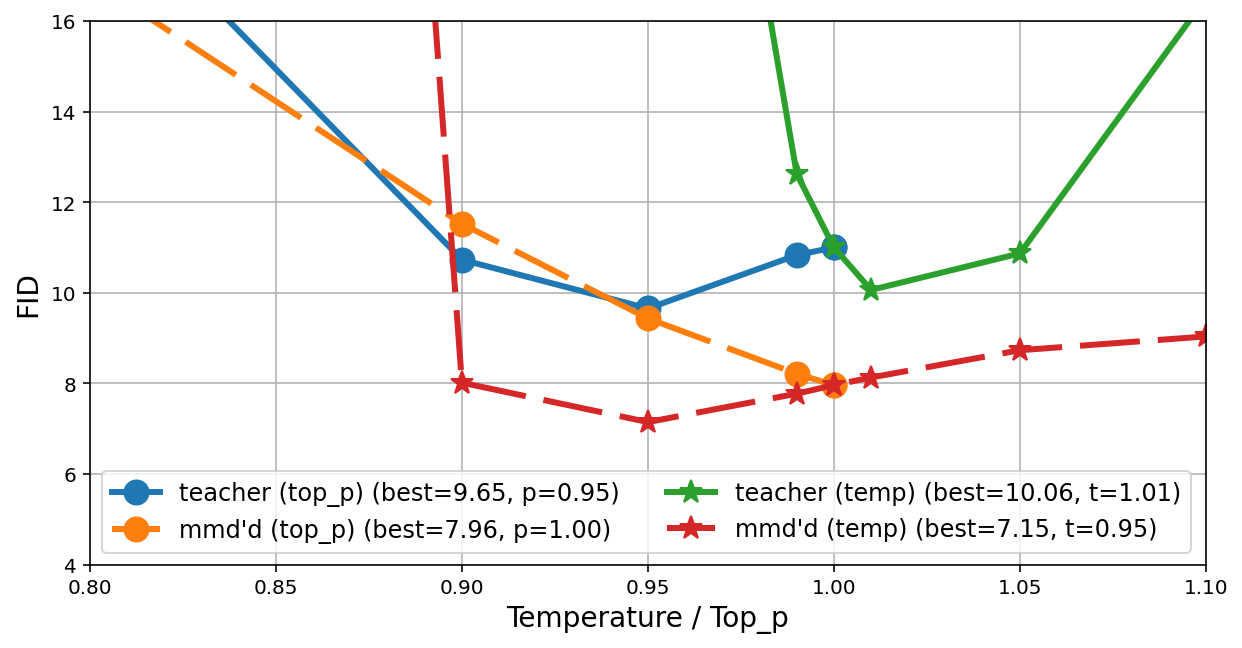}
        \caption{Transformer Uniform architecture, temperature and top $p$ sampling.}
    \end{subfigure}
    \caption{FID performance vs sampling temperature or top $p$ value in posterior sampling.}
    \label{fig:eval_posterior_sampling_settings}
\end{figure*}

\paragraph{Posterior Sampling Settings}
See \autoref{fig:eval_posterior_sampling_settings}.

We evaluate two ways to adjust the posterior sampling during evaluation time.
\begin{enumerate}
    \item Temperature scaling, by adding:\newline\fcolorbox{blue}{white}{
{\footnotesize
\texttt{x\_sample = jnp.argmax(x\_logits + self.sampling\_temperature * g, axis=-1)
}}}
    \item Top P sampling, by using a selection mechanism to use only the top $p$ percent of the probability mass.
\end{enumerate}

\begin{figure*}[h]
    \begin{subfigure}[b]{.475\textwidth}
        \centering
        \includegraphics[width=\textwidth]{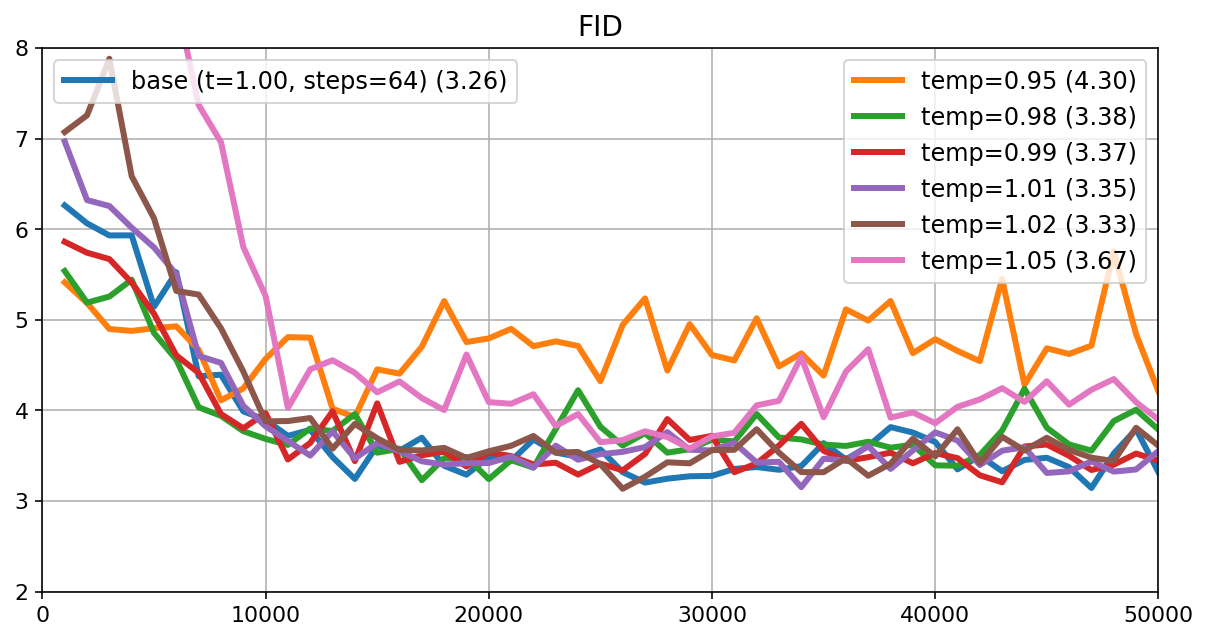}
        \caption{Masked diffusion}
    \end{subfigure}
    \begin{subfigure}[b]{.4755\textwidth}
        \centering
        \includegraphics[width=\textwidth]{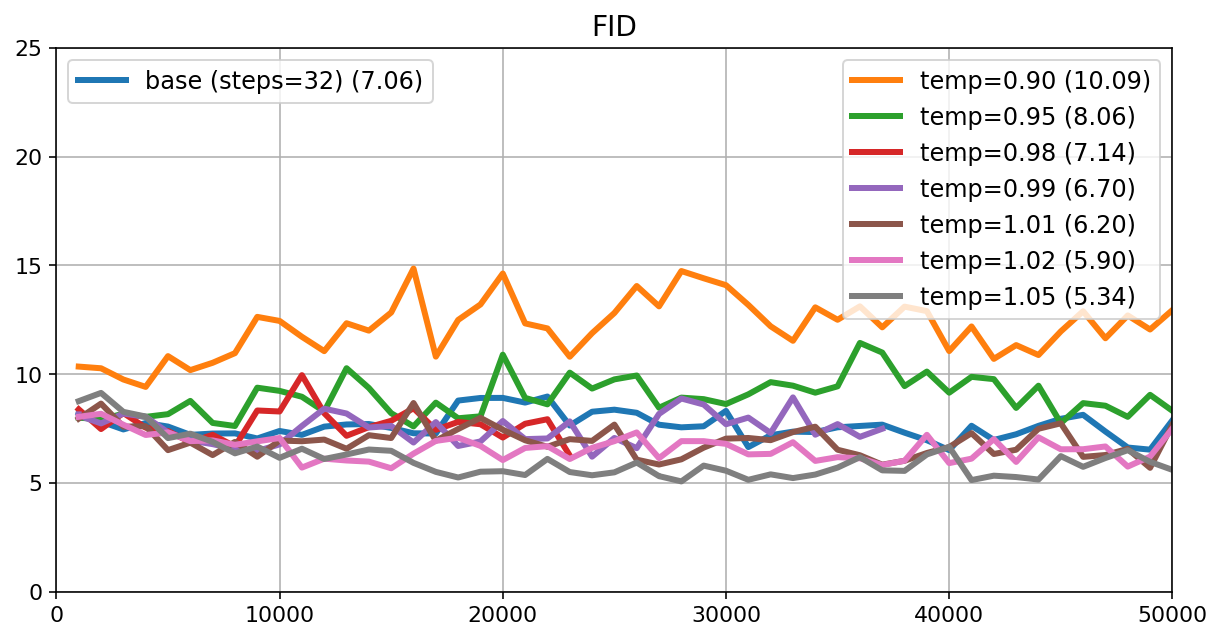}
        \caption{Transformer Uniform}
    \end{subfigure}
    \caption{FID performance vs teacher temperature while MMD'ing the student model.}
    \label{fig:mmd_teacher_temperature}
\end{figure*}

\paragraph{MMD'ing with teacher temperature}
See \autoref{fig:mmd_teacher_temperature}

\begin{figure*}[h]
    \begin{subfigure}[b]{.475\textwidth}
        \centering
        \includegraphics[width=\textwidth]{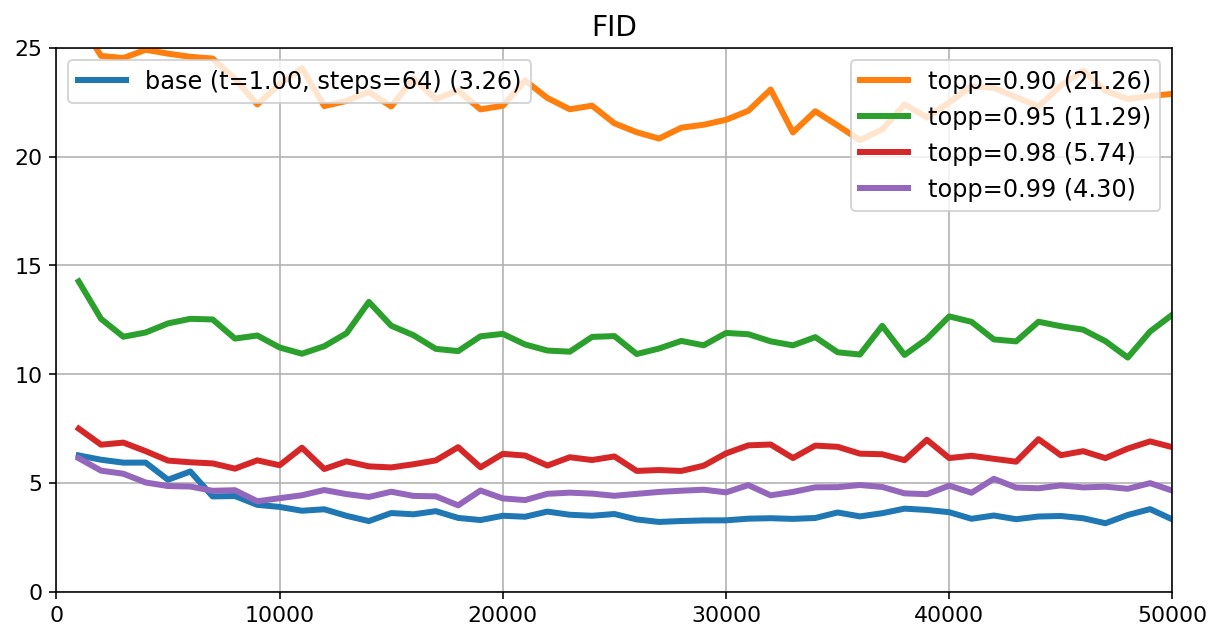}
        \caption{Masked}
    \end{subfigure}
    \begin{subfigure}[b]{.4755\textwidth}
        \centering
        \includegraphics[width=\textwidth]{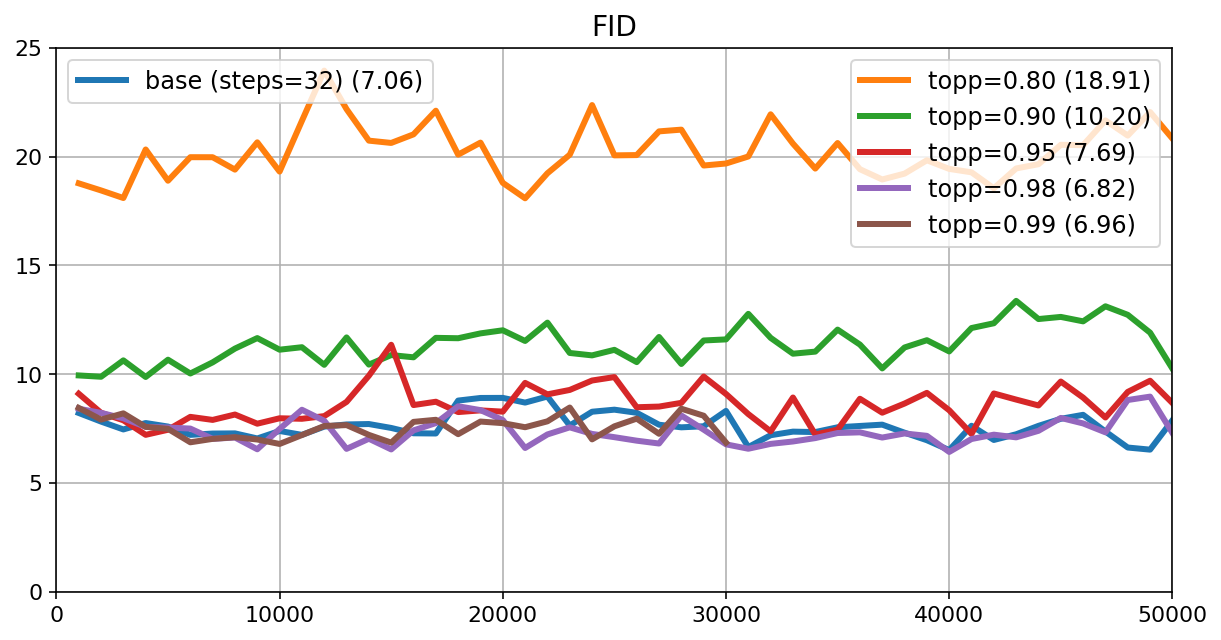}
        \caption{Uniform}
    \end{subfigure}
    \caption{FID performance vs teacher top $p$ value while MMD'ing the student model.}
    \label{fig:mmd_teacher_topp}
\end{figure*}

\paragraph{MMD'ing with teacher top $p$ sampling}
See \autoref{fig:mmd_teacher_topp}

\section{D-MMD for other discrete diffusion models}
\label{app:other_diffusions}
In certain discrete diffusion models, it is (surprisingly) not always true that $E_q[x|z_t]$ is the optimal solution for $x_\theta(z_t)$. An example is the case of uniform diffusion as parametrized in \citep{hoogeboom2021argmaxflows,austin2021structured}. We will discuss how one could do D-MMD for parametrizations such as these.

\paragraph{Background Discrete Diffusion}
It is helpful to study the simplified posterior parametrization as it covers uniform diffusion and any other discrete process that interpolates from data $x$ to a factorized stationary distribution $\pi$ so that $z_t = \mathrm{Cat}\left(z_s | \alpha_t x + (1 - \alpha_t) \pi\right)$. In that case the posterior of this process given $x$ equals \citep{Sahoo2024simpleandeffective}:
\begin{equation}
    q(z_s|z_t,x)=\text{Cat}\left(z_s| \frac{[\alpha_{t|s}z_t+(1-\alpha_{t|s})1\pi^\top z_t]\odot[\alpha_s x+(1-\alpha_s)\pi]}{\alpha_t z_t^\top x+(1-\alpha_t)z_t^\top\pi}\right) = \text{Cat}(z_s | \pi_{z_s}(x, z_t)),
\end{equation}
for which we define the shorthand probability vector $\pi_{z_s}(x, z_t)$. As a result, writing the loss component for discrete diffusion simplifies to $L_t(x, \hat{x}_\theta(z_t), z_t) = \mathrm{KL}\left(\pi_{z_s}(x, z_t) || \pi_{z_s}(\hat{x}_\theta(z_t), z_t) \right)dt^{-1}$ where $s = t - dt$. One can either simply choose a discretization for which $dt > 0$ or take the limit $dt \to 0$ which requires some subsequent algebraic manipulation.

In these cases recall that $L_t(x, \hat{x}_\theta(z_t), z_t) = \mathrm{KL}\left(\pi_{z_s}(x, z_t) || \pi_{z_s}(\hat{x}_\theta(z_t), z_t) \right)dt^{-1}$. In this case the subtraction of the two KL terms cancels out the negative entropy term $\sum \pi_{s-ds}(\hat{x}_\eta) \log \pi_{s-ds}(\hat{x}_\eta)$ leading to the loss:  
\begin{equation}
     \mathcal{L}_{\text{D-MMD}}(\eta) = L_s(\hat{x}_\eta, \hat{x}_\theta(z_s), z_s) -  L_s(\hat{x}_\eta, \hat{x}_\phi(z_s), z_s) = \sum_c \pi_{s-ds}(\hat{x}_\eta)_c \left( \log \pi_{s-ds}(\hat{x}_\phi(z_s)) - \log \pi_{s-ds}(\hat{x}_\theta(z_s))\right)_c,
\end{equation}
A fixed point for this algorithm occurs when $g_\eta(x)$ is distributed as the data (approximated by teacher) distribution $q_\theta(x)$, in which case $\hat{x}_\phi(z_s) = \hat{x}_\theta(z_s)$ and both the generator and the auxiliary model have an update of zero.

For the auxiliary model, flipping signs and ignoring constants the loss can be written as:
\begin{align}
     \mathcal{L}_{\text{AUX}}(\phi) &= L_s(\hat{x}_\eta, \hat{x}_\phi(z_s), z_s) + L_s(\hat{x}_\theta(z_s), \hat{x}_\phi(z_s), z_s) \\
     &\stackrel{c}{=} CE(\pi_{s-ds}(\hat{x}_\eta) | \pi_{s-ds}(\hat{x}_\phi(z_s))) + CE(\pi_{s-ds}(\hat{x}_\theta(z_s)) | \pi_{s-ds}(\hat{x}_\phi(z_s)))
\end{align}
This has the optimum $\pi_{s-ds}(\hat{x}_\phi(z_s)) = \frac{1}{2}\left(\mathbb{E}_{g_\eta(x)}[\pi_{s-ds}(x)] + \pi_{s-ds}(\hat{x}_\theta(z_s))\right)$. As a result, when $g_\eta$ is distributed as the data (or the approximation of the teacher) the optimum is $\pi_{s-ds}(\hat{x}_\phi(z_s)) = \pi_{s-ds}(\hat{x}_\theta(z_s))$.
\end{document}